%% file: main-arxiv-submission.tex
\documentclass[10pt,twocolumn,letterpaper]{article}

\usepackage{cvpr}
\usepackage{times}
\usepackage{epsfig}
\usepackage{graphicx}
\usepackage{amsmath}
\usepackage{amssymb}
\usepackage[normalem]{ulem}
\usepackage{color}

\usepackage{dsfont}

\usepackage{bbding}
\usepackage{ulem}

\usepackage{array,multirow,textcomp}
\newcolumntype{C}[1]{>{\centering\let\newline\\\arraybackslash\hspace{0pt}}m{#1}}
\newcommand{\thickhat}[1]{\mathbf{\hat{\text{$#1$}}}}

\usepackage[pagebackref=true,breaklinks=true,letterpaper=true,colorlinks,bookmarks=false]{hyperref}

\cvprfinalcopy 

\def\cvprPaperID{4728} 

\begin{document}


\title{Domain Agnostic Feature Learning for \\ Image and Video Based Face Anti-spoofing}
\author{Suman Saha\\
ETH Zurich\\
{\tt\small suman.saha@vision.ee.ethz.ch}
\and
Wenhao Xu\\
ETH Zurich\\
{\tt\small wenhxu@student.ethz.ch}
\and
Menelaos Kanakis\\
ETH Zurich\\
{\tt\small menelaos.kanakis@vision.ee.ethz.ch}
\and
Stamatios Georgoulis\\
ETH Zurich\\
{\tt\small stamatios.georgoulis@vision.ee.ethz.ch}
\and
Yuhua Chen\\
ETH Zurich\\
{\tt\small yuhua.chen@vision.ee.ethz.ch}
\and
Danda Pani Paudel\\
ETH Zurich\\
{\tt\small paudel@vision.ee.ethz.ch}
\and
Luc Van Gool \\
KU Leuven \& ETH Zurich\\
{\tt\small vangool@vision.ee.ethz.ch}}

\maketitle

\input{text/abstract}

\input{text/intro}
\input{text/relwork}

\input{text/approach}

\input{text/exp}

\input{text/exp_suman}
\input{text/conclusion}
\\
\\
\input{text/appendix-v2}


{\small
\bibliographystyle{ieee_fullname}
\bibliography{egbib}
}

\end{document}

%% file: text/abstract.tex
\begin{abstract}

Nowadays, the increasingly growing number of mobile and computing devices has led to a demand for safer user authentication systems. Face anti-spoofing is a measure towards this direction for biometric user authentication, and in particular face recognition, that tries to prevent spoof attacks. The state-of-the-art anti-spoofing techniques leverage the ability of deep neural networks to learn discriminative features, based on cues from the training set images or video samples, in an effort to detect spoof attacks. However, due to the particular nature of the problem, i.e. large variability due to factors like different backgrounds, lighting conditions, camera resolutions, spoof materials, etc., these techniques typically fail to generalize to new samples. In this paper, we explicitly tackle this problem and propose a class-conditional domain discriminator module, that, coupled with a gradient reversal layer, tries to generate live and spoof features that are discriminative, but at the same time robust against the aforementioned variability factors. Extensive experimental analysis shows the effectiveness of the proposed method over existing image- and video-based anti-spoofing techniques, both in terms of numerical improvement as well as when visualizing the learned features.   

\end{abstract}

%% file: text/intro.tex
\section{Introduction}

\begin{figure}[t]
\begin{center}
  \includegraphics[width=0.9\linewidth]{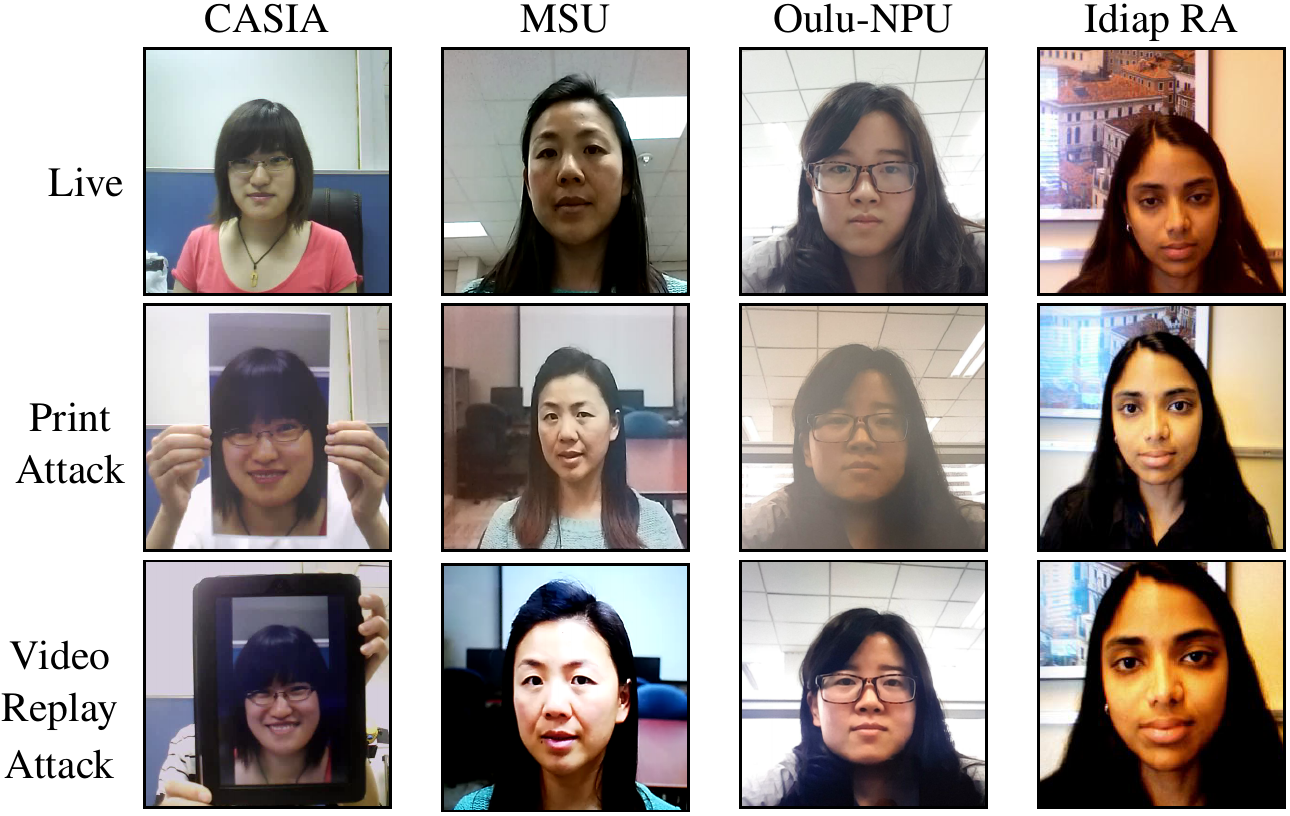}
\end{center}
\vskip -0.3cm
  \caption{
  Sample frames from the four publicly available face anti-spoofing datasets:
   CASIA-MFSD~\cite{zhang2012face},
   MSU-MFSD~\cite{wen2015face},
   Oulu-NPU~\cite{boulkenafet2017oulu} and
   Idiap Replay-Attack (RA)~\cite{chingovska2012effectiveness}.
   Note that, a large variability can be observed due to factors like different attack instruments, backgrounds, lighting conditions, camera resolutions etc.
   resulting significant domain shift among these datasets.
  }
  \vskip -0.5cm
\label{fig:fas_datasets}

\end{figure}

Increasingly, people use computing devices, such as laptops and smartphones, to work, pay their bills, purchase things as well as interact with their social circle, entertain themselves, \etc Given the constant use we make of these devices, it is important to develop convenient, yet secure, ways to log into them. Lately, biometric authentication, and in particular face recognition, has emerged as an attractive way of user identification due to the unique nature of each individual's face in combination with the ease-of-use of this approach (e.g. Apple's FaceID). At the same time, however, hackers have become more inventive in their attempts to spoof someone's face in order to fool the authentication system. Typical examples include printing one's face picture on paper (print attack), playing a video depicting the person's face on another device (replay attack), wearing a special mask to closely imitate someone's facial appearance (mask attack), \etc Understandably, being able to effectively detect such attacks, formally known as face anti-spoofing (FAS), is a critical problem in computer vision.

On the one hand, traditional approaches to face anti-spoofing rely on hand-crafted features, like LBP~\cite{chingovska2012effectiveness,de2012lbp,de2013can}, HoG~\cite{komulainen2013context,yang2013face} and SURF~\cite{boulkenafet2017face}, to detect differences in texture between the live and spoof face images, or heuristics, like eyeblink~\cite{pan2007eyeblink}, lip motion~\cite{kollreider2007real}, and visual rhythms~\cite{pinto2015using}, to identify regularities that are absent from the spoof attacks. However, the aforementioned methods are either not applicable to all possible spoof attacks, \ie print, replay, and mask, or they fail to generalize to different datasets, since the learned features specialize to the `trained' textures, which largely vary between datasets due to factors like different backgrounds, lighting conditions, camera resolutions, spoof materials, \etc as can be seen in Fig. \ref{fig:fas_datasets}.

On the other hand, modern approaches use convolutional neural networks (CNNs)~\cite{yang2014learn,li2016original,patel2016cross,atoum2017face,liu2018learning,jourabloo2018face} that have shown impressive performance in many computer vision tasks, largely attributed to the great representational power of their learned features when trained on large-scale datasets. Despite the improved performance, there are still open challenges in FAS. A notable one is the domain\footnote{The term domain in this paper is used to refer to a dataset.} shift~\cite{li2018unsupervised} problem. The latter occurs when a network trained on one dataset (source domain) is tested on a completely unseen dataset (target domain). This is referred to as ``cross-testing'' in the FAS literature, while training and testing on the same dataset is referred to as ``intra-testing''. The existing deep learning based approaches show promising results for intra-testing, but their performance dramatically degrades when evaluated under a cross-testing setup~\cite{shao2019multi}. The main reason for this performance drop is the feature distribution dissimilarity (see Fig. \ref{fig:domain_shift}) between the source and target domains caused by several dataset specific cues, such as differences in: (1) environmental conditions (illumination, background), (2) spoofing mediums (printers, display screens), and (3) the quality of video capturing devices (different mobile phones, tablets). Thus, a model learns to differentiate between live and spoof samples based on these dataset dependent cues, but fails to correctly classify samples from unknown datasets having different sets of cues.

In this paper, we address the aforementioned domain shift problem in FAS under the domain generalization setting. That is, the network is trained on multiple datasets (source domains), but then tested on a completely unseen dataset (target domain). Our goal is to generate domain agnostic feature representations using the source domain samples that would generalize to the unseen target domain samples, so that each sample, regardless of its domain origin, can effectively be classified as live or spoof. To this end, we propose the use of class-conditional domain discriminator modules coupled with a gradient reversal layer \cite{ganin2014unsupervised}. The former take the feature representations generated from a backbone network, and try to classify from which source domain each sample comes, conditioned on the class it belongs (i.e. live or spoof). The latter acts as an identity transform during the forward propagation, but multiplies the gradient by a certain negative constant during the backward propagation, essentially reversing the objective of its subsequent layers. In our case, this practically means that the backbone network is now tasked with the extra objective of generating live and spoof feature representations that are indistinguishable across domains. Note that, our method works for both image-based and video-based inputs, but we explicitly avoid to include extra components as input, like depth~\cite{atoum2017face} or rPPG signals~\cite{liu2018learning}, as the latter would require expensive ground truth labels in order to train the network.  

Our key contributions can be summarized as:
(1) a class-conditional domain discriminator module (\S~\ref{subsec:ccdg}) which coupled with a gradient reversal layer promotes the learning of domain agnostic features;
(2) an LSTM network (\S~\ref{subsec:approach:overview},\ref{para:dg_resnet_opt_cost}) to learn temporal domain agnostic features as complementary information;
(3) state-of-the-art results on the four challenging domain generalization test sets (\S~\ref{subsec:comparison_to_sota}) with an accompanying visual analysis of the feature embedding (\S~\ref{subsec:exp:tsne}) and class activation maps (\S~\ref{subsec:exp:clsactmap}). 


%% file: text/relwork.tex
\section{Related work}

\begin{figure*}[ht]
\begin{center}
   \includegraphics[width=0.8\linewidth]{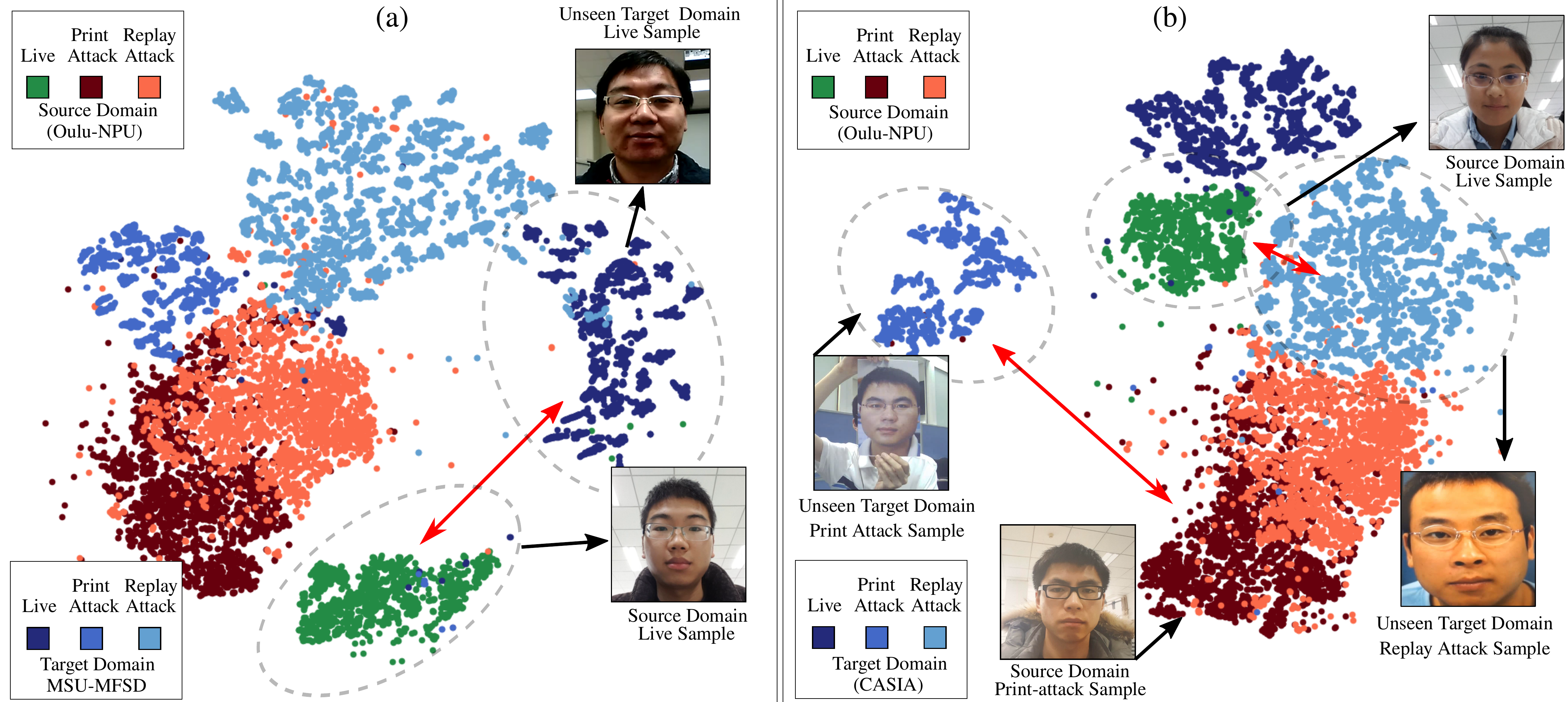} 
\end{center}
\caption{
A t-SNE visualization of CNN features from a ResNet50 backbone trained on multiple source domains (i.e. FAS datasets) and tested on an unseen target domain. For better visualization we show only one source and one target domain in these plots. We can easily recognize the inherent domain shift problem in face anti-spoofing. That is, the live and spoof samples from the source and target domains are not properly aligned in the feature space, resulting in poor generalization of the learned feature representations on the target domain.
}
\label{fig:domain_shift}
\end{figure*}

In what follows, we describe traditional, feature-based as well as modern, CNN-based approaches to FAS.
We then elaborate on the few domain generalization works on FAS.

\textbf{Traditional approaches}
Before the advent of CNNs~\cite{krizhevsky2012imagenet}, typical approaches to face anti-spoofing combined the use of hand-crafted features with shallow classification techniques to detect differences in texture between the live and spoof images. The most characteristic examples of hand-crafted features include LBP~\cite{maatta2011face,chingovska2012effectiveness,de2012lbp,de2013can}, HoG~\cite{komulainen2013context,yang2013face}, DoG~\cite{tan2010face,peixoto2011face}, SIFT~\cite{patel2016secure}, and SURF~\cite{boulkenafet2017face}. As for the classifier, these works rely on SVM and, to a lesser degree, on LDA. In a similar vein, other traditional approaches employed heuristics to leverage `liveliness' cues that are not present in a spoof attack. Examples of such heuristics are eyeblink~\cite{pan2007eyeblink,sun2007blinking}, lip motion~\cite{kollreider2007real}, visual rhythms~\cite{pinto2015using}, Haralick texture features~\cite{agarwal2016face}, audio~\cite{chetty2006audio,chetty2010biometric}, dynamic textures~\cite{de2014face}, or others~\cite{tan2010face,liu20163d,liu2018remote}. Another way to address face anti-spoofing is to make use of temporal cues in videos, for example, Bharadwaj et al.~\cite{bharadwaj2014face} tried to use motion magnification to enhance facial expressions, while ~\cite{komulainen2013complementary} tried using motion and texture cues. To account for factors of variation in the compared images, such as illumination, pose, \etc, the use of different color spaces (HSV, YCbCr)~\cite{boulkenafet2015face,boulkenafet2016face}, image distortion analysis~\cite{wen2015face}, or a transformation to the temporal domain~\cite{bao2009liveness,siddiqui2016face} and Fourier spectrum~\cite{li2004live}, have been explored. In general, these traditional methods are either not applicable to all possible spoof attacks, \ie print, replay, mask, or they fail to generalize to different datasets, since the learned features specialize to the 'trained' textures, which largely vary between datasets due to factors of variation like different backgrounds, lighting conditions, camera resolutions, spoof materials, \etc

\textbf{CNN-based approaches}
The impressive results achieved by applying CNNs to the tasks of image classification and object recognition~\cite{krizhevsky2012imagenet,girshick2014rich,simonyan2014very,szegedy2015going} motivated researchers to employ them to other computer vision tasks too. Face anti-spoofing is no exception. The obvious choice is to replace the hand-crafted features with features learned from generic CNNs - known for their great representational power when trained on large-scale datasets - which is what was done in~\cite{patel2016cross,li2016original}. Differently, Yang~\etal~\cite{yang2014learn} used a CNN as a binary classifier, to assign live/spoof labels to the input images. They used a multi-scale pyramid of the RGB images as input, whereas Feng~\etal~\cite{feng2016integration} explored the use of multiple cues, such as image quality and motion cues. Next, Xu~\etal~\cite{xu2015learning} incorporated video inputs and proposed an LSTM-CNN model to take advantage of the information from the extra frames. Dynamic textures were proposed in~\cite{shao2017deep,shao2018joint} to extract different facial motions. Recently, Atoum~\etal~\cite{atoum2017face} introduced a multitasking-inspired approach that combines the estimation of texture and depth features for binary live/spoof classification, which was later extended by Liu~\etal~\cite{liu2018learning} to also include fusion with temporal supervision, \ie rPPG signals. Finally, Joorabloo~\etal~\cite{jourabloo2018face} followed a different path and inversely decomposed a spoof face into a spoof noise and a live face using a GAN architecture, and consequently utilized the spoof noise for classification.
Bresan et al.~\cite{bresan2019facespoof} explore depth, salience and illumination maps associated with a pre-trained CNN for FAS.
They use combination of source domains (i.e. NUAA \cite{tan2010face}, Idiap Replay-Attack~\cite{chingovska2012effectiveness}, CASIA-MFSD~\cite{zhang2012face} dataset),
different from ours, and thus their method is not directly comparable.

The aforementioned works, despite showing improved performance, partially attributed to the use of CNNs, still face open challenges when it comes to generalizing across domains (i.e. datasets). As mentioned, there is an inherent domain shift~\cite{li2018unsupervised} between the different FAS datasets (\eg Replay Attack~\cite{chingovska2012effectiveness} and CASIA-FASD~\cite{zhang2012face}), which in turn leads to poor cross-testing results. In this paper, we go beyond current CNN-based approaches and explicitly tackle the domain shift problem in FAS without relying on supervision from extra cues, like depth or rPPG signals, that would require a significant annotation effort to acquire.

\textbf{Domain generalization approaches.}
To tackle the domain shift problem across different datasets, domain adaptation~\cite{gopalan2011domain,fernando2013unsupervised,tzeng2014deep,ganin2014unsupervised,ganin2016domain} and generalization~\cite{khosla2012undoing,muandet2013domain,xu2014exploiting,ghifary2015domain,ghifary2017scatter,li2017deeper,motiian2017unified,li2018domain} techniques have been used in computer vision. The goal in each case is to bridge the distribution gap between data from source and target domains in order to create domain agnostic feature representations that generalize to new domains. In this paper, we are mostly interested in domain generalization techniques, which have been largely unexploited in FAS, with the following exceptions.
Li~\etal~\cite{li2018learning} encouraged the learning of generalized feature representations by taking both spatial and temporal information into consideration and minimizing a cross-entropy loss together with a generalization loss.
Tu~\etal~\cite{tu2019learning} proposed the use of Total Pairwise Confusion loss for CNN training in conjunction with a Fast Domain Adaptation component into the CNN model to account for domain changes.
Shao~\etal~\cite{shao2019multi} combined the learning a generalized feature space that is shared by multiple discriminative source domains with dual-force triplet mining constraint to improve the discriminability of the learned feature space.
In general, compared to the aforementioned works our framework offers better integration to multiple domains, and, as will be shown in Sec.~\ref{sec:exp}, achieves significantly improved results on four public datasets.

%% file: text/approach.tex
\section{Proposed Approach} \label{sec:method}

\begin{figure*}[t]
\begin{center}
   \includegraphics[width=0.99\linewidth]{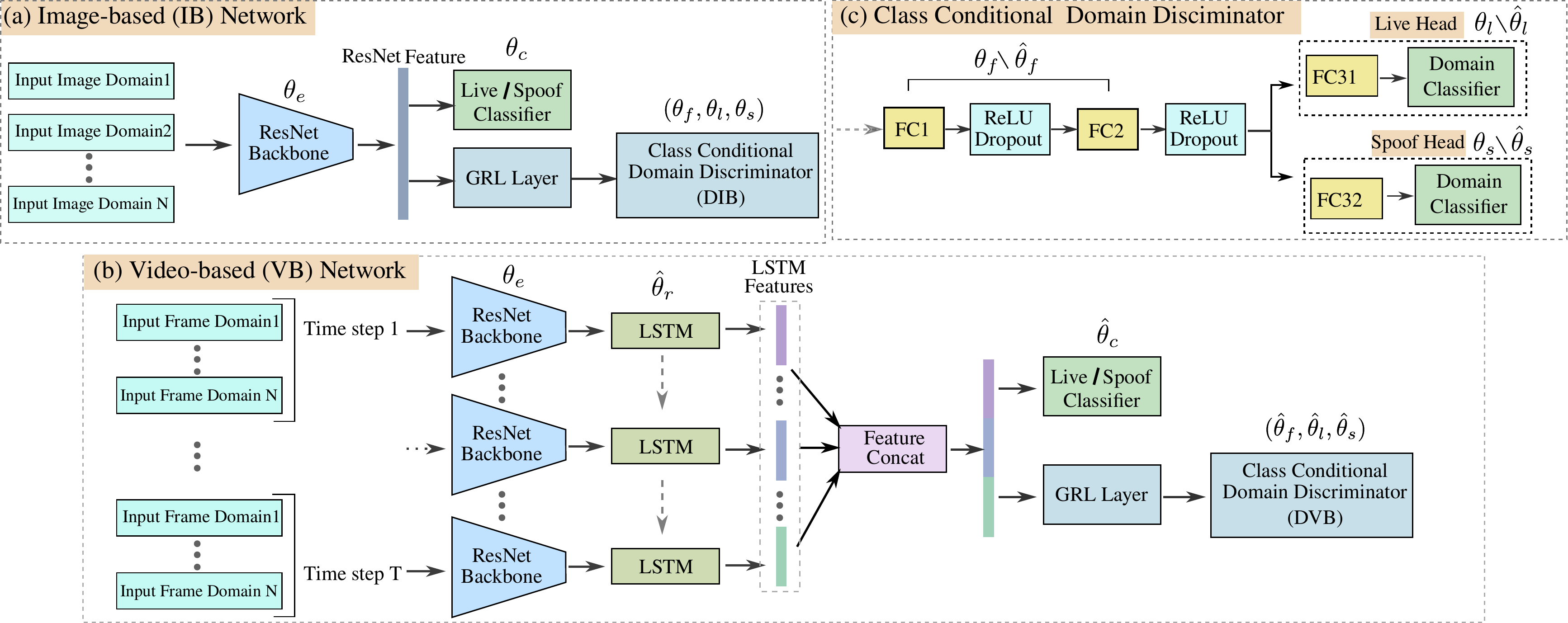} 
\end{center}
   \caption{Overview of the different components of the proposed approach. See Section~\ref{sec:method} for more details.}
\label{fig:overview}
\end{figure*}

\subsection{The domain shift problem in face anti-spoofing}
Our main goal is to learn generalized feature representations in order to address the domain shift problem that inherently exists among FAS datasets. That is, the distribution dissimilarities between live and spoof samples that belong to multiple source and unseen target domains. To illustrate this problem, we use t-SNE plots (Fig.~\ref{fig:domain_shift}) generated from the CNN features of a ResNet50~\cite{he2016deep} backbone trained on multiple source domains (i.e. FAS datasets) for live/spoof classification, and tested on an unseen target domain. As can be seen in Fig.~\ref{fig:domain_shift} (a), the CNN features of the live samples from the unseen target domain are far away from the live samples of the source domain in the feature space. Similarly in Fig.~\ref{fig:domain_shift} (b), we can see that the print attack features from the target domain are far apart from the source domain's print attacks, and the target domain's replay attack features are shifted towards the live samples of the source domain. It is quiet evident from these illustrations that even deep neural networks, like ResNet models, are not sufficient on their own to tackle the problem. This calls for dedicated mechanisms that can leverage the common attributes shared across multiple source domains to learn more generic feature representations. The term common attributes is used here to refer to the common intrinsic properties of the print and replay attacks across multiple domains. For example, although these attacks might have been generated using different spoofing mediums (i.e. different printers or video capturing devices), or under different environmental conditions (e.g. illumination, background scene), they are inherently based on paper materials or display screens. Thus, by leveraging these common attributes one could expect that better feature representations can be learned from the shared and discriminative information across multiple source domains, that is robust for live/spoofing classification and at the same time domain agnostic. We expect such representations to demonstrate better generalization on unseen target domains.

\subsection{System overview} \label{subsec:approach:overview}

To tackle the aforementioned problem, we propose a novel framework which learns both image- and video-based domain agnostic feature representations (see Fig. \ref{fig:overview}). More specifically, a ResNet backbone (encoder) is trained to minimize the live/spoof classification loss, while at the same time it competes against a class-conditional domain discriminator (\S \ref{subsec:ccdg}) coupled with a gradient reversal layer to maximize the domain classification loss of live and spoof samples respectively. During the training process the encoder gradually learns the shared and discriminative feature representations. A system overview is given in Fig.~\ref{fig:overview}.

You can observe two variations (see Fig. \ref{fig:overview}). First, an image based (IB) network that follows an image-level training, in which a training example consists of an image and its associated ground-truth label (either ``live'' or ``spoof''). This is to demonstrate the scenario where only a single image is given as input, and the system has to decide if this is a spoof attack or not. However, FAS can also be a video classification problem, i.e. we expect the final output to be a live/spoof label for an input video sample. Thus, a CNN trained following an image-level protocol might fail if we process the results on a frame-by-frame basis, as the video itself usually contains richer information. For such instances, we want the network to learn strong temporal features which are complementary to the spatial representation learned by the IB network. Based on this idea, we also propose a video-based (VB) network which is trained along-side the IB network, following an alternating training scheme~\cite{liu2018learning}.
This VB network uses the same ResNet backbone, i.e. model parameters of the ResNet backbone are shared between the IB and VB networks. Unlike the IB network, the VB network inputs video sequence and processes these through multiple long-short term memory (LSTM) units and outputs a single class label for each input video sequence.

\subsection{Class-conditional domain discriminator} 
\label{subsec:ccdg}
In Fig.~\ref{fig:overview} (c), we show the network architecture of our proposed class-conditional domain discriminator (CCDD). CCDD consists of two fully connected layers, FC1 and FC2, followed by a \emph{live} and a \emph{spoof} head. FC1 and FC2 layers are followed by a ReLU and a dropout layer. During training, an SGD mini-batch that consists of live and spoof training examples is processed through the FC1 and FC2 layers. Consequently, the outputs of the FC2 layer are first split into ``live'' and ``spoof'' batches, and then, they are passed as input to their respective heads. The \emph{live} and \emph{spoof} heads have the same layer configuration, i.e. each consists of a single linear transformation layer followed by a domain classifier. They output two score vectors $\mathbf{s}_{l}$ and $\mathbf{s}_{f}$ having $D$ scores, \ie the softmax probability scores for each domain. Note that, we use the same network architecture for the image- and video-based CCDD (DIB and DVB in Fig.\ref{fig:overview} (a) \& (b)).

The proposed CCDD coupled with the gradient reversal layer imposes the desired conditional invariance property on the learned feature representations. The conditional invariance is realized by the class-conditional losses (see below), which consider the source domain label information only and aim to make the representation in each class indistinguishable across domains. We present a t-SNE visualization (\S \ref{subsec:exp:tsne}) to demonstrate that the proposed CCDD learns to correctly align the live and spoof features of the target domain with the features of source domains. Besides, we present quantitative experimental results to attest the effectiveness of the CCDD. 
A more detailed network design is provided in \ref{appendix:detailed_network_design}. 

\subsection{Gradient reversal layer}  
\label{para:grl}
The gradient reversal layer (GRL)~\cite{ganin2014unsupervised} was originally proposed for unsupervised domain adaptation. Instead, we couple CCDD with GRL in order to learn domain agnostic features from multiple source domains for FAS. In particular, we use two GRL layers, one in the image-based and another one in the video-based network (Fig.\ref{fig:overview}). What GRL essentially does, is to reverse the gradient by multiplying it by a negative scalar (i.e. the adaptation factor $\lambda_{GRL}$) during the backward propagation. During the forward propagation, it leaves the input unchanged, \ie it acts as an identity transform. By doing so, it essentially reverses the objective of its subsequent layers, i.e. CCDD in our case. What this practically means, is that the backbone network is now tasked with the extra objective of generating live and spoof feature representations that are indistinguishable across multiple source domains.

\begin{table*}[ht]
\footnotesize
\centering
\caption{Comparison to state-of-the-art FAS methods on four domain generalization test sets.}
\scalebox{0.9}{
\begin{tabular}{c||cc|cc|cc|ccc}
    \hline
 \multirow{2}{4em}{Method}  & \multicolumn{2}{c|}{O\&C\&I$\to$M}  & \multicolumn{2}{c|}{O\&M\&I$\to$C} & \multicolumn{2}{c|}{O\&C\&M$\to$I}  & \multicolumn{2}{c}{I\&C\&M$\to$O}\\ 
                            & HTER(\%)      & AUC(\%)           & HTER(\%)          & AUC(\%)       & HTER(\%)          & AUC(\%)           & HTER(\%)          & AUC(\%)   \\ \hline \hline
                            
MS LBP \cite{maatta2011face} & 29.76         & 78.50             & 54.28                  &  44.98             &  50.30                 &  51.64                 &  50.29                 &   49.31        \\ \hline
Binary CNN \cite{yang2014learn} & 29.25              &   82.87                &  34.88                 &  71.94             &   34.47                &   65.88                &  29.61                 &  77.54         \\ \hline
IDA \cite{wen2015face}    &  66.67             & 27.86                 &    55.17              &   39.05            &  28.35                 &   78.25                &  54.20                 &   44.59        \\ \hline
Color Texture \cite{boulkenafet2016face}     &  28.09             &  78.47                &    30.58               &  76.89             &   40.40                &   62.78                &   63.59                &  32.71         \\ \hline
LBPTOP \cite{de2014face}      &   36.90            &  70.80                 &   42.60                &  61.05             &   49.45                &  49.54                 &  53.15                 &  44.09         \\   \hline     
Auxiliary(Depth Only) \cite{liu2018learning}        &   22.72            & 85.88                  & 33.52                 &  73.15             &  29.14                 &  71.69                &   30.17                &  77.61         \\ \hline
Auxiliary(All)  \cite{liu2018learning}            &     -          &      -             &        28.4           &     -          &      27.6             &         -          &           -        &      -     \\ \hline 
Ours   &  \textbf{15.42}            &    \textbf{91.13}                 &     \textbf{17.41}                &     \textbf{90.12}            &    \textbf{15.87}                &     \textbf{91.72}              & \textbf{14.72}                  &  \textbf{93.08}         \\ \hline 
\end{tabular}
}
\label{table:sota_comp}
\end{table*}

\begin{table*}[ht]
\footnotesize
\centering
\caption{Comparison to state-of-the-art domain generalization FAS methods on four domain generalization test sets.}
\scalebox{0.9}{
\begin{tabular}{c||cc|cc|cc|ccc}
    \hline
 \multirow{2}{4em}{Method}  & \multicolumn{2}{c|}{O\&C\&I$\to$M}  & \multicolumn{2}{c|}{O\&M\&I$\to$C} & \multicolumn{2}{c|}{O\&C\&M$\to$I}  & \multicolumn{2}{c}{I\&C\&M$\to$O}\\ 
                            & HTER(\%)      & AUC(\%)           & HTER(\%)          & AUC(\%)       & HTER(\%)          & AUC(\%)           & HTER(\%)          & AUC(\%)   \\ \hline \hline
                            
MMD-AAE \cite{li2018domain}  &   27.08            &  83.19                 &   44.59                &   58.29            &  31.58                 &   75.18                &   40.98                &   63.08            \\ \hline 
MADDG \cite{shao2019multi}                    &   17.69            &   88.06                &    24.5               &    84.51           &       22.19            &    84.99               &     27.98              &     80.02      \\ \hline 
Ours                        &  \textbf{15.42}            &    \textbf{91.13}                 &     \textbf{17.41}                &     \textbf{90.12}            &    \textbf{15.87}                &     \textbf{91.72}              & \textbf{14.72}                  &  \textbf{93.08}         \\ \hline 
\end{tabular}
}
\label{table:dg_sota_comp}
\end{table*}

\subsection{Optimization cost}  
\label{para:dg_resnet_opt_cost}
First, we specify the energy function used to optimize the IB network (Fig.\ref{fig:overview} (a)). Consider the following notations: $\theta_{f}$, $\theta_{l}$ and $\theta_{s}$ be the model parameters of the common layers (i.e. FC1 \& FC2), \emph{live} and \emph{spoof} heads of the DIB respectively; $\theta_{e}$ and $\theta_{c}$ be the model parameters of the encoder (i.e. the ResNet backbone) and the label classifier (i.e. the live/spoof classifier); $L_{l}$ and $L_{s}$ be the domain classification losses (\ie multinomial) for the \emph{live} and \emph{spoof} heads that penalize for incorrect domain label prediction separately for the ``live'' and ``spoof'' training examples; $L_{c}$ be the label classification (e.g. multinomial) loss that penalizes for incorrect class label (i.e. ``live'' or ``spoof'') prediction; $i$ denotes the index for a training example and $F$ be the number of training examples, \ie $i = \{1, 2, \dots, F\}$; $b_{i}$ be a binary variable denoting the class label of the $i$-th example, \ie $b_{i}=0$ indicates that the example is live and $b_{i}=1$ that it is a spoof.
During the IB network training, the encoder's model parameters $\theta_{e}$ learn to minimize the discrepancy in the class conditional distribution \cite{li2018deep} across different domains. This is done by maximizing the domain classification losses of the \emph{live} and \emph{spoof} heads of the DIB. In other words, it tries to make the feature distributions (belonging to a class $c \in C$) maximally similar across different domains. At the same time, the \emph{live} and \emph{spoof} heads seek parameters $\theta_{l}$ and $\theta_{s}$ which minimize the class conditional domain classification losses. This yields as energy function for our IB network:
\vskip -0.5cm
\begin{align} \label{eq:imnet}
    & E(\theta_{e},\theta_{c}, \theta_{f}, \theta_{l}, \theta_{s}) =  \sum_{i=1\dots F} L^{i}_{c}(\theta_{e},\theta_{c}) \nonumber \\
    &   + \lambda_{IB} \Bigg( \sum_{\substack{i=1\dots F \\ b=0}} L^{i}_{l}(\theta_{e},\theta_{f}, \theta_{l})   +  \sum_{\substack{i=1\dots F \\ b=1}} L^{i}_{s}(\theta_{e},\theta_{f}, \theta_{s})\Bigg)
\end{align}
\vskip -0.4cm
Now, we specify the energy function used to optimize the VB network (Fig.\ref{fig:overview} (b)).
Let: $\thickhat{\theta}_{r}$ be the model parameters of the LSTM network;
$\thickhat{\theta}_{f}$, $\thickhat{\theta}_{l}$ and $\thickhat{\theta}_{s}$ be the model parameters of the common layers (i.e. FC1 \& FC2), \emph{live} and \emph{spoof} heads of the video-based class-conditional domain discriminator respectively; $\thickhat{\theta}_{c}$ be the model parameters of the LSTM's label classifier (i.e. the live/spoof classifier).
In a similar fashion, during the VB network training the encoder's and LSTM's model parameters (i.e. $\theta_{e}$ and $\thickhat{\theta}_{r}$) learn to minimize the discrepancy in the class conditional distribution across different domains by maximizing the domain classification losses of the \emph{live} and \emph{spoof} heads of the DVB. At the same time, the \emph{live} and \emph{spoof} heads seek parameters $\thickhat{\theta}_{l}$ and $\thickhat{\theta}_{s}$ which minimize the class conditional domain classification losses. This yields as energy function for our VB network:
\vskip -0.5cm
\begin{align}  \label{eq:vmnet}
    & E(\theta_{e}, \thickhat{\theta}_{r}, \thickhat{\theta}_{c}, \thickhat{\theta}_{f}, \thickhat{\theta}_{l}, \thickhat{\theta}_{s}) =  \sum_{i=1\dots F} \thickhat{L}^{i}_{c}(\theta_{e}, \thickhat{\theta}_{r}, \thickhat{\theta}_{c}) \nonumber \\
    &   + \lambda_{VB} \Bigg( \sum_{\substack{i=1\dots F \\ b=0}} \thickhat{L}^{i}_{l}(\theta_{e}, \thickhat{\theta}_{r}, \thickhat{\theta}_{f}, \thickhat{\theta}_{l})   +  \sum_{\substack{i=1\dots F \\ b=1}} \thickhat{L}^{i}_{s}(\theta_{e}, \thickhat{\theta}_{r},\thickhat{\theta}_{f}, \thickhat{\theta}_{s})\Bigg)
\end{align}

\vskip -0.5cm
$\thickhat{L}^{i}_{c}$,  $\thickhat{L}^{i}_{l}$ and $\thickhat{L}^{i}_{s}$ are the
live/spoof classification loss and the domain classification losses (for the live and spoof heads) for VB network.
$\lambda_{IB}$ and $\lambda_{VB}$ are the scalar parameters weighting the relative importance of the two loss terms in Eq. \ref{eq:imnet} and Eq. \ref{eq:vmnet} respectively.
Note that, the encoder's model parameters $\theta_{e}$ are shared across the image- and video-based networks.

%% file: text/exp.tex
\section{Experiments} \label{sec:exp}

\begin{figure*}[ht]
\begin{center}
   \includegraphics[width=0.9\linewidth]{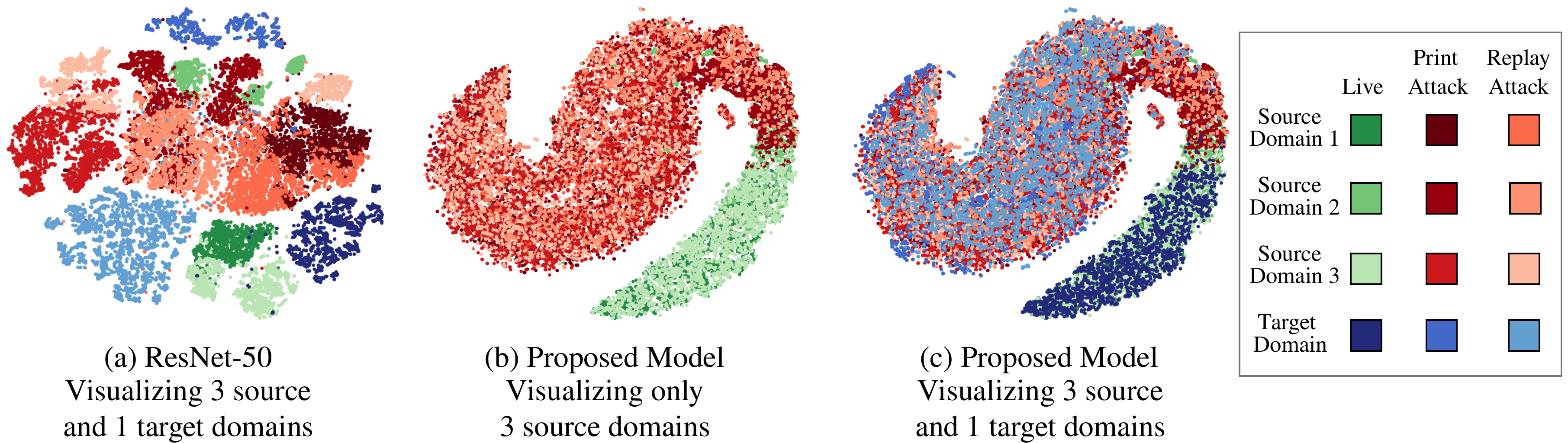} 
\end{center}
   \caption{
  A t-SNE plot of the CNN features coming from ResNet (a) vs our full model (b,c), both trained on three source domains and tested on an unseen target domain (best viewed in color). 
  Note that, the live features of source and target domains are far apart  (a);
  similar trend can be noticed for the spoof features of source and target domain,
  but our model learns to group together all live and spoof features (from multiple source domains) into two different clusters (b), thus improving the classification accuracy.
  Importantly, the learned representations generalize well on the target domain (c).
   }
\label{fig:tsne}
\end{figure*}

\begin{table*}[ht]
\footnotesize
\centering
\caption{An ablation study of the different components in the proposed FAS architecture on four domain generalization test sets.}
\scalebox{0.88}{
\begin{tabular}{cccc||cc|cc|cc|ccc}
    \hline
 \multirow{2}{*}{ResNet} & \multirow{2}{*}{DIB} & \multirow{2}{*}{LSTM} & \multirow{2}{*}{DVB}  & \multicolumn{2}{c|}{O\&C\&I$\to$M}         & \multicolumn{2}{c|}{O\&M\&I$\to$C}         & \multicolumn{2}{c|}{O\&C\&M$\to$I}                 & \multicolumn{2}{c}{I\&C\&M$\to$O}\\ 
 &    &    &   & HTER(\%)      & AUC(\%)                   & HTER(\%)          & AUC(\%)               & HTER(\%)          & AUC(\%)                       & HTER(\%)          & AUC(\%)   \\ \hline \hline

\checkmark & & & & 21.66              & 89.64              & 25.92               &  82.16              & 20.12             &  90.1                        & 18.81          & 89.53 \\ \hline 
\checkmark & \checkmark & &  & 18.33             & 90.58               & 21.29                & 85.82              & 17.63             & 86.3                          & 17.05         & 90.01 \\ \hline 
\checkmark & & \checkmark &  & 17.92             & 90.27               & 19.26               & 87.85                & 18.0              & 89.78                         & 16.42         & 90.82  \\ \hline 
\checkmark & & \checkmark & \checkmark & 18.33             & 88.25               & 21.11               & 88.22                 & 18.25             & 85.61                         & 17.05         & 91.09  \\ \hline 
\checkmark & \checkmark & \checkmark &  & \textbf{14.58}     & \textbf{92.58}          & 18.7                & 89.35              & \textbf{15.13}    & \textbf{95.76}               & 14.86         &  93.00        \\ \hline
\checkmark & \checkmark & \checkmark & \checkmark & 15.42             & 91.13               & \textbf{17.41}      & \textbf{90.12}       & 15.87             & 91.87                         & \textbf{14.72}  & \textbf{93.08} \\ \hline

\end{tabular}
}
\label{table:ablation}
\end{table*}




\subsection{Experimental setting}
\textbf{Datasets.}
We evaluate our method on four publicly available FAS datasets: Oulu-NPU~\cite{boulkenafet2017oulu} (O for short), CASIA-MFSD~\cite{zhang2012face} (C for short), Idiap Replay-Attack~\cite{chingovska2012effectiveness} (I for short), and MSU-MFSD~\cite{wen2015face} (M for short). 

\textbf{Training and evaluation.}
We consider a dataset to be one domain in our experiments. Our model learns domain generalized representations from three out of four datasets, as in~\cite{shao2019multi}. In particular, we randomly select three datasets as the source domains, and the remaining unseen domain, which is not accessed during training, is kept for evaluation only. Half Total Error Rate (HTER)~\cite{bengio2004statistical} and Area Under Curve (AUC) are used as the evaluation metrics in our experiments. 

\textbf{Implementation details.}
We use ResNet-50~\cite{he2016deep} as our backbone network. The dimension of the input image is $224 \times 224$. During training, we use SGD optimizer, and follow an alternative training approach~\cite{liu2018learning} to train both our IB and VB networks (Fig.~\ref{fig:overview}). We use a constant learning rate of $0.0003$, momentum $0.9$ and weight decay $0.00001$. The mini-batch size for the IB network is $48$, i.e. $16$ training images from each of the three domains. For the VB network, the mini-batch size is $6$, i.e. $2$ training video sequences from each of the three source domains, and the LSTM sequence length is $8$. The LSTM's input dimension is $2048$, while the hidden layer dimension is $256$. We use a constant GRL adaptation factor ($\lambda_{GRL} = -0.2$)~\cite{ganin2014unsupervised}, and set the $\lambda_{IB}$ and $\lambda_{VB}$ to $1$. 
Additional experimental details are presented in \ref{appendix:add_exp_details}.


\subsection{Comparison to the state-of-the-art} 
\label{subsec:comparison_to_sota}

In Table~\ref{table:sota_comp}, we compare our full model against state-of-the-art FAS methods. Our proposed method outperforms~\cite{maatta2011face,yang2014learn,wen2015face,boulkenafet2016face,de2014face,liu2018learning} on all the four domain generalization test sets. The significantly better performance mostly lies in the ability to learn rich generalizable features, which adapt well to the unseen target domain (see Fig.\ref{fig:tsne}). Note that, these FAS methods do not explicitly address the domain shift problem, and thus naturally fail to generalize well on unseen target domains. In contrast, our proposed method explicitly learns a generalizable representation by leveraging the available information (live and spoof examples with ground truth labels) from multiple source domains. In particular, it learns to map all the live and spoof samples (from multiple source domains) to a common feature space where the live and spoof features are far apart, while being domain invariant at the same time.

In addition, we compare against the state-of-the-art domain generalization FAS method~\cite{shao2019multi} 
and also compare to the related state-of-the-art method in domain generalization for the face anti-spoofing task: MMD-AAE \cite{li2018domain} as in \cite{shao2019multi}.
These methods explicitly address the domain shift problem. 
Table~\ref{table:dg_sota_comp} shows this comparison, where our method consistently achieves much better performance. We conclude that the proposed method can overcome the distribution dissimilarities in the feature space more effectively. Moreover,~\cite{shao2019multi} is relatively expensive and not end-to-end trainable, in contrast to our method.

\subsection{Ablation study on model components} 
\label{subsec:exp:abl}

So far, we have shown results with our \textit{full model} that contains all the different components, 
i.e. ResNet backbone (ResNet), 
image-level domain discriminator (DIB), 
LSTM module (LSTM), 
and video-level domain discriminator (DVB). 
In what follows, we present a detailed ablation study when using different combinations of these components.
The experimental results on all four domain generalization test sets are summarized in Table~\ref{table:ablation}. 
When we mention DIB or DVB in Table~\ref{table:ablation}, it automatically includes the associated GRL layer.

To demonstrate the applicability of the proposed model components, we first setup our own baseline for the ablation study. 
The baseline is comprised of a ResNet-50 backbone and a live/spoof classifier which is trained on the four different domain generalization training sets.
Our baseline itself exhibits some desirable performance.
In the supplementary material, we report experiments with a lighter ResNet backbone.
When adding DIB on top of the ResNet backbone, the results are consistently improved on all four test sets.
Additionally adding LSTM, the results are again improved significantly. 
Finally, our full model boosts the results further.
Combining ResNet and LSTM, provides slightly better results on three test setups
compared to the model using ResNet and DIB. 
However, adding DVB to the model with ResNet and LSTM does not bring any further improvements.
However, when DVB is jointly trained with ResNet, DIB and LSTM,  i.e. our full model, improves over the ResNet baseline.
This observation verifies that by exploiting both spatial (DIB or image-based) and temporal (DVB or video-based) domain-agnostic features 
our proposed model can achieve the best results on the two most challenging domain generalization test sets (O\&M\&I$\to$C and I\&C\&M$\to$O).




\subsection{Visualization of the learned CNN features} 
\label{subsec:exp:tsne}
Fig.~\ref{fig:tsne} depicts t-SNE plots of the CNN activations (\ie features) coming from our ResNet baseline vs our full model.
Both networks were trained on 3 source domains (\ie Oulu-NPU, CASIA-MFSD and MSU-MFSD) and tested on a target domain (\ie Idiap  replay-attack). Note that, the plots in (b) and (c) are generated using the same trained model, i.e. our full model, and the same set of live and spoof samples. For the sake of better visualization, however, we have deactivated the visualization of the target domain in (b). As can be seen in (b), our model learns more discriminative features for live and spoof images. What is more interesting is that the representation learned by our model aligns well with unseen target domain's live and spoof features, as can be seen by activating the target domain visualization in (c). In contrast, the ResNet learnt representation shows relatively weaker generalization ability on the target domain, as shown in (a). In the latter case, the live, print- and replay-attack features from multiple source domains are far apart in the feature space, whereas our model learns to minimize this inter-domain distances between live and spoof features, as shown in (b, c). From these visualizations, we can conclude that our network generalizes well on the target domain. Particularly observe in (c) how the target domain live and spoof features are properly aligned with the live and spoof features of the source domains in (b).

\begin{figure}[ht]
\begin{center}
   \includegraphics[width=0.99\linewidth]{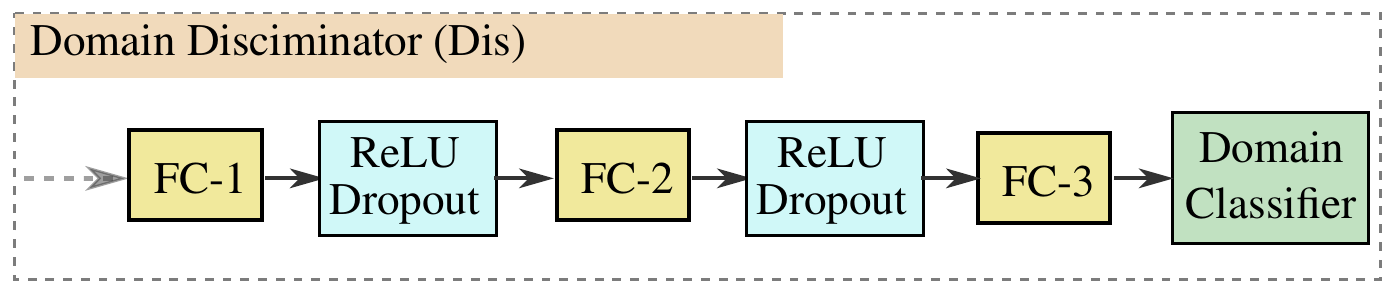}
\end{center}
   \caption{Architectural components of our default Domain Discriminator network (Dis).
   }
\label{fig:dp_net_arch}
\end{figure}
\vskip -0.5cm
\begin{figure}[ht]
\begin{center}
   \includegraphics[width=0.99\linewidth]{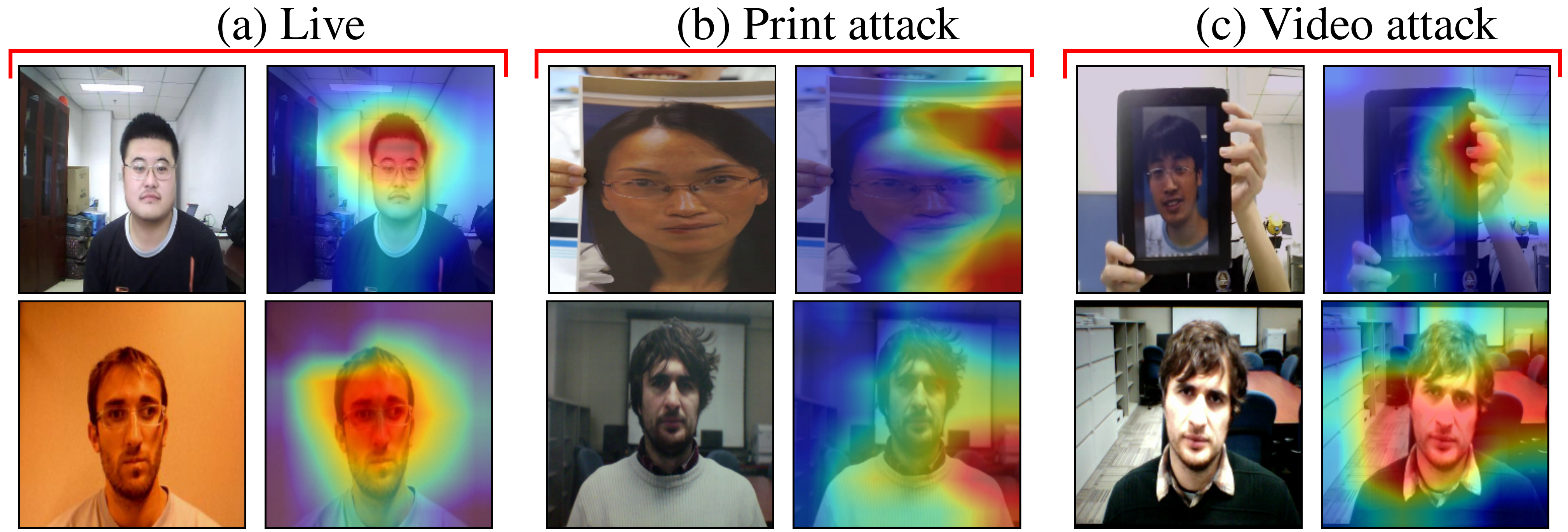} 
\end{center}
   \caption{Activation map visualization of the proposed network. 
   For each column (a), (b) and (c), the original input images and its associated network class activation maps are shown.}
\label{fig:gradcam_visual}
\end{figure}

\vskip -0.5cm
\begin{table}[ht]
\footnotesize
\centering
\caption{Performance comparison of different domain discriminators on three domain generalization test sets.}
\scalebox{0.9}{
\begin{tabular}{ccc||c|c|c}
    \hline
\multirow{3}{*}{ResNet} &\multirow{3}{*}{Dis} &\multirow{3}{*}{DIB}                 & S\&O\&I\&R      & S\&O\&C\&R      & S\&C\&I\&R   \\ 

 \multicolumn{3}{c||}{}& $\to$C      & $\to$I      & $\to$O   \\ 

\multicolumn{3}{c||}{} & HTER(\%)    & HTER(\%)    & ACER(\%) \\ \hline \hline
\checkmark & & & 17.5                              & 20.6                        & 10.27  \\ \hline
\checkmark&\checkmark & & 15.3                              & 17.7                        & 8.75 \\     \hline
\checkmark&\checkmark &\checkmark & 15.1                              & 17.0                        & 23.4  \\  \hline
\checkmark& &\checkmark & \textbf{14.0}                     & \textbf{14.7}                & \textbf{8.05}  \\ \hline
\end{tabular}
}
\label{table:impact_of_diff_domain_disc}
\end{table}

\subsection{Impact of different domain discriminators} \label{subsec:impact_diff_dd}
We conduct experiments to analyze the effect of using different domain discriminators on the FAS performance. We consider two domain discriminator architectures: the proposed DIB (Fig. \ref{fig:overview}), and the default domain classifier  (Fig. \ref{fig:dp_net_arch}) originally proposed by~\cite{ganin2014unsupervised}
for unsupervised domain adaptation (Dis in Table~\ref{table:impact_of_diff_domain_disc}).
Note that, for the experiments in this section we used -- only for training purposes -- two more datasets, i.e. SiW (S for short)~\cite{liu2018learning} and Idiap replay-mobile (R for short)~\cite{costa2016replay}. 
Following~\cite{liu2018learning}, when testing on Oulu-NPU dataset, we use the ACER metric.
From Table~\ref{table:impact_of_diff_domain_disc}, it can be seen that our ResNet-DIB gives the best performance. When ResNet-Dis is used, the performance degrades slightly. Even combining Dis with DIB degrades the performance heavily on Oulu-NPU. From these experiments, we observe that learning feature representations from multiple source domains conditioned on class labels (\ie live and spoof) can provide discriminative and domain agnostic features, while conditioning them on domain labels only may not correctly align the live and spoof features, resulting poor classification accuracy. As the proposed DIB has access to both class (live and spoof) and domain labels, in contrast to Dis, it is able to learn better representations by correctly grouping live features from multiple source domain into one cluster and spoof features into another (see Fig. \ref{fig:tsne}).

\subsection{Class activation map visualization} 
\label{subsec:exp:clsactmap}
In this section, we provide a visual analysis of the class activation maps to get an intuition about the decisions the network makes when making a particular prediction. For this visualization, we use the Grad-CAM~\cite{selvaraju2017grad} technique. In Fig. \ref{fig:gradcam_visual} we show the class activation maps for the live, print and replay attack test samples. Some interesting observations can be made. The network gives more importance to the facial regions for detecting a ``live'' class (see Fig. \ref{fig:gradcam_visual} (a)) which is intuitive as most of the information about a live face comes from the facial region. For example, the texture of a live skin, the eye blinking, head motion etc. On the other hand, for print attacks the network pays more attention to the surface of the paper (on which the face image is printed) (Fig. \ref{fig:gradcam_visual} (b)). For video replay attacks, if strong features like ``a hand in the background'' and 
``a tablet screen'' are present then the network takes decision from these salient information (Fig. \ref{fig:gradcam_visual} (c) top). In the absence of such strong features, it tries to see both the facial region and the background (Fig. \ref{fig:gradcam_visual} (c) bottom).

\vskip -1cm

%% file: text/conclusion.tex
\section{Conclusion}

In this paper, we addressed an inherent problem in face anti-spoofing, i.e. the large variability in factors such as the different backgrounds, lighting conditions, camera resolutions, spoof materials, etc., makes feature representations learned by CNNs for this task too domain-dependent, leading to decreased performance when testing on unseen domains. We propose a solution based on generalizable feature learning that naturally fits this 'domain shift' problem in both image-based and video-based face anti-spoofing. We provide extensive experimentation on multiple aspects of our approach, and among others, we demonstrate state-of-the-art performance across different test sets, we illustrate the qualitative improvement of the learned feature representations w.r.t. generalization, and visualize through the class activation maps the network's attention when making predictions. For future work, we would like to use multi-modal inputs and apply domain agnostic multi-modal feature learning to further improve the classification accuracy.

%% file: text/appendix-v2.tex
\appendix

\section{Appendix}

\begin{table*}[h!t]
\footnotesize
\centering
\caption{The architectural details of the proposed network.}
\scalebox{0.99}{
\begin{tabular}{lll|lll}
\hline
\multicolumn{3}{c|}{CCDD (Class-conditional Domain Discriminator)} & \multicolumn{3}{c}{Live/Spoof Classifier}  \\
Layer & Input Dim. & Output Dim. & Layer & Input Dim. & Output Dim. \\ \hline \hline
FC1         & 2048 & 1024 & FC1 & 2048 & 512 \\  
ReLU        &      &      & ReLU &     &     \\
Dropout     &      &      & Dropout &     &     \\

FC2         & 1024 & 1024 & FC2 & 512 & 2 (num. class labels) \\ 
ReLU        &      &      &  &     &     \\
Dropout     &      &      &  &     &     \\

FC31 (Live Head)         & 1024 & 3 (num. source domains) &  &  &  \\ 
FC32 (Spoof Head)         & 1024 & 3 (num. source domains) &  &  &  \\ 
\hline
\end{tabular}
}
\label{table:network_arch}
\end{table*}

\subsection{Detailed network design} \label{appendix:detailed_network_design}
In this section, we present a detailed architectural design of the proposed network.
We use the ResNet-50 as our backbone network and PyTorch for implementation purpose.
In Table~\ref{table:network_arch}, we show the layer-wise network design of our proposed Class-conditional Domain Discriminator (CCDD) and the live/spoof classifier (LSC).
Note that, the ResNet-50 backbone inputs a $224 \times 224$ image and outputs a $2048$ dimensional feature vector which is passed as the input to both the CCDD and LSC.
The FC31 and FC32 layers (i.e. the live and spoof heads) of CCDD output $3$ softmax probability scores for the $3$ source domains.
The FC2 (or final) layer of the LSC outputs $2$ softmax probability scores for the $2$ class labels - ``live'' and ``spoof''.

The shape of the input tensor to the LSTM network is $[T \times B \times 2048]$ where $T$ is the sequence length  and $B$ is the SGD mini-batch size.
We set $T$ and $B$ to $8$ and $2$ respectively.
For each time step $t$, the LSTM outputs $256$ dimensional feature vector, where $t = 1,2,...,8$.
These $8$ feature vectors are concatenated to a single feature vector of dim. $2048$ which is then passed as input to the LSC.

\begin{table*}[h!t]
\footnotesize
\centering
\caption{Domain generalization training, validation and test sets used in this work.}
\scalebox{0.99}{
\begin{tabular}{c|c|c|c}
\hline
Dataset Name        &  Training Set                             & Validation Set                                & Test Set \\ \hline \hline
O\&C\&I$\to$M       & Training sets from Oulu-NPU,              & Validation sets from Oulu-NPU,                & MSU-MFSD test set \\ 
                    & CASIA-MFSD and Idiap Replay-Attack.       & CASIA-MFSD and Idiap Replay-Attack.             &   \\ \hline
O\&M\&I$\to$C       & Training sets from Oulu-NPU,              & Validation sets from Oulu-NPU,                 & CASIA-MFSD test set \\ 
                   & MSU-MFSD and Idiap Replay-Attack.         & MSU-MFSD and Idiap Replay-Attack.                &  \\ \hline

O\&C\&M$\to$I       & Training sets from Oulu-NPU,              & Validation sets from Oulu-NPU,                &  Idiap Replay-Attack test set \\ 
                   & CASIA-MFSD and MSU-MFSD.                  & CASIA-MFSD and MSU-MFSD.                       &   \\ \hline
I\&C\&M$\to$O       & Training sets from Idiap Replay-Attack,   & Validation sets from Idiap Replay-Attack,      &  Oulu-NPU test set \\ 
                    & CASIA-MFSD and MSU-MFSD.                   & CASIA-MFSD and MSU-MFSD.                      &   \\ \hline
\end{tabular}
}
\label{table:dg_dataset}
\end{table*}

\subsection{Additional experimental details} \label{appendix:add_exp_details}
We use four domain generalization datasets as in~\cite{shao2019multi} which are created from the following publicly available face anti-spoofing datasets:
Oulu-NPU~\cite{boulkenafet2017oulu} (O for short), CASIA-MFSD~\cite{zhang2012face} (C for short), Idiap Replay-Attack~\cite{chingovska2012effectiveness} (I for short), and MSU-MFSD~\cite{wen2015face} (M for short).
In Table~\ref{table:dg_dataset}, we present the training, validation and test set details for each of these four datasets.
CASIA-MFSD and MSU-MFSD don't have validation sets and following the standard practice, we use a subset of the training set as the validation set for both of these datasets.
At inference time, we receive the predictions from both image- and video-based live/spoof classifiers (see Fig.3 in the main paper).
As a final output, we select the one which gives the best performance on the validation set.
We initialize the ResNet-50 backbone with ImageNet \cite{krizhevsky2012imagenet} pretrained weights. 

\begin{table*}[h!t]
\footnotesize
\centering
\caption{Face anti-spoofing performance (HTER\%) improvements with smaller backbone network (ResNet-18). }
\scalebox{0.99}{
\begin{tabular}{l||c|c|c|c}
    \hline
Model                                              & O\&C\&I$\to$M       & O\&M\&I$\to$C            & O\&C\&M$\to$I                 &   I\&C\&M$\to$O  \\ \hline  \hline  
ResNet-18 backbone                                 & 27.5                &  31.67                   & 21.63                         &  14.83   \\
\textbf{Our model} (uses ResNet-18 backbone)       & \textbf{22.5}       &  \textbf{28.52}          &  \textbf{20.38}             &  \textbf{12.78}       \\  \hline  
\end{tabular}
}
\label{table:lighter_backbones}
\end{table*}

\subsection{Experimental results with smaller backbone} \label{appendix:exp_with_smaller_backbone}
In this section, we present our experimental results with a smaller backbone (i.e. ResNet-18) compared to the ResNet-50 used in the main paper.
We replace the ResNet-50 backbone in our proposed framework with ResNet-18 and train the model.
In Table~\ref{table:lighter_backbones}, we show that even if our proposed framework uses a weaker backbone network,
it shows consistent improvements on the four challenging domain generalization test sets.
Note that, \emph{I\&C\&M $\to$O} has the smallest training set
among these four domain generalization datasets (Table \ref{table:dg_dataset}).
For this smaller dataset, our proposed framework achieves better performance with a ResNet-18 backbone.

\begin{table*}[h!t]
\footnotesize
\centering
\caption{Assessing the model's generalizability on three domain generalization test sets.}
\scalebox{0.99}{
\begin{tabular}{c|c|c|c}
    \hline
Model                 & S\&O\&I\&R $\to$C  HTER(\%)     & S\&O\&C\&R $\to$I   HTER(\%)    & S\&C\&I\&R $\to$O  ACER(\%) \\   \hline \hline
ResNet-50               & 17.5                              & 20.6                        & 10.27  \\ 
\textbf{Our IB Network}  & \textbf{14.0}                     & \textbf{14.7}                & \textbf{8.05}  \\ \hline
\end{tabular}
}
\label{table:improving_dg}
\end{table*}

\subsection{Assessing the model's generalizability} 
\label{subsec:improving_dg}
To assess the model's generalization ability, we increase the number of source domains. 
By doing so, we allow the network to see live and spoof examples with large variations in subjects, environmental conditions, attack instruments, video capturing devices, etc.
One may argue that in this case the improvements merely come by adding more data.
To ensure this is not the case, we also compare results against the ResNet baseline on the same data.
As shown in Table~\ref{table:improving_dg}, our imaged based network (DIB) achieves consistent improvements over the ResNet baseline, on three different domain generalization test settings. 
Note that, for the experiments in this section we used -- only for training purposes -- two more datasets, i.e. SiW (S for short)~\cite{liu2018learning} and Idiap replay-mobile (R for short)~\cite{costa2016replay}. 
Although, further improvements can be achieved by adding the video based network (LSTM and DVB), here we demonstrate improvements only in image-based FAS and exclude the video-based case. 
Following~\cite{liu2018learning}, when testing on Oulu-NPU dataset, we use the ACER metric and report results (in Table~\ref{table:improving_dg}) by averaging the ACERs over the four test protocols.

\begin{table*}[h!t]
\footnotesize
\centering
\caption{Comparison to the existing domain adaptation based face anti-spoofing methods on four domain adaptation test sets.}
\scalebox{0.99}{
\begin{tabular}{c||c|c|c|c}
\hline
\multirow{2}{4em}{Method}                       &  {M$\to$I}            & {I$\to$M}              & {I$\to$C}                      & {C$\to$I}\\ 
                                                & HTER(\%)              & HTER(\%)              & HTER(\%)                      & HTER(\%)       \\ \hline \hline
Li et al. \cite{li2018unsupervised}	            & 33.30	                & 33.20	                & \textbf{12.30}	            & 39.20  \\ \hline
Tu et al. \cite{tu2019deep}	                    & 27.50	                & 25.83	                & -	                            & -  \\  \hline
Tu et al. \cite{tu2019learning}                 & 25.80                 & 23.50	                & 23.50	                        & 21.40  \\ \hline
\textbf{Our full model}                                          & \textbf{9.38}	        & \textbf{12.91}	    & 16.11	                        &\textbf{11.38} \\ \hline 

\end{tabular}
}
\label{table:da_exp}
\end{table*}

\subsection{Domain adaptation experiments}
In this section, we compare the face anti-spoofing performance of our proposed approach with the existing domain adaptation based FAS methods~\cite{li2018unsupervised,tu2019deep,tu2019learning}.
For these experiments, we follow standard unsupervised domain adaptation training setting, \ie the network is trained using examples from a single source domain (with ground-truth labels) and unlabelled training examples from the target domain. 
To align with the domain adaptation training setup, we use the default domain discriminator (see Section 4.5, Fig.5).
The results are presented in Table~\ref{table:da_exp}.
Out of four domain adaptation test sets, our proposed framework outperforms~\cite{li2018unsupervised,tu2019deep,tu2019learning} on the three test sets, and shows comparable results on the remaining one.
These results demonstrate that our proposed model can be exploited under both domain adaptation and domain generalization settings. 
Note that, in this paper we are interested in the latter setting, yet as we observe from these experimental results, our model with small adaptations can achieve significant improvements for the former setting too.

Li~\etal~\cite{li2018unsupervised} have different strategies for domain adaptation and we select whatever gives out the best result for their model and compare to ours in Table~\ref{table:da_exp}.
Li~\etal~\cite{li2018learning} approach is not comparable to ours because they assume that different domains are just different camera models, which is quite restricting. 
In our case they are different datasets, allowing us to also address changes in spoofing mediums, illumination and background.

\begin{figure*}[h!t]
\begin{center}
  \includegraphics[width=0.8\linewidth]{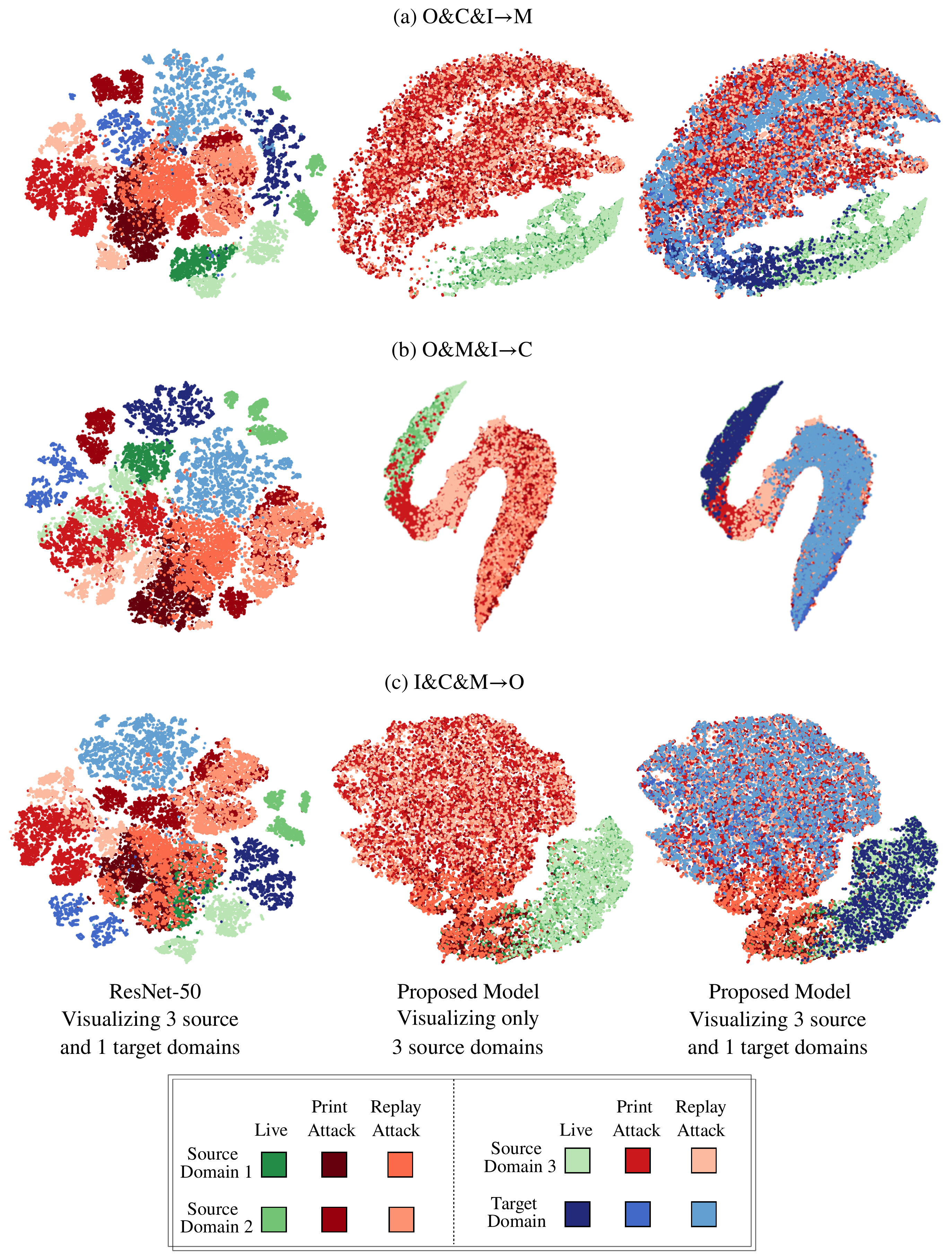} 
\end{center}
  \caption{A t-SNE Visualization of the learned CNN features coming from ResNet-50 baseline and from our proposed network.
  }
\label{fig:tsne2}
\end{figure*}

\begin{figure*}[h!t]
\begin{center}
  \includegraphics[width=0.9\linewidth]{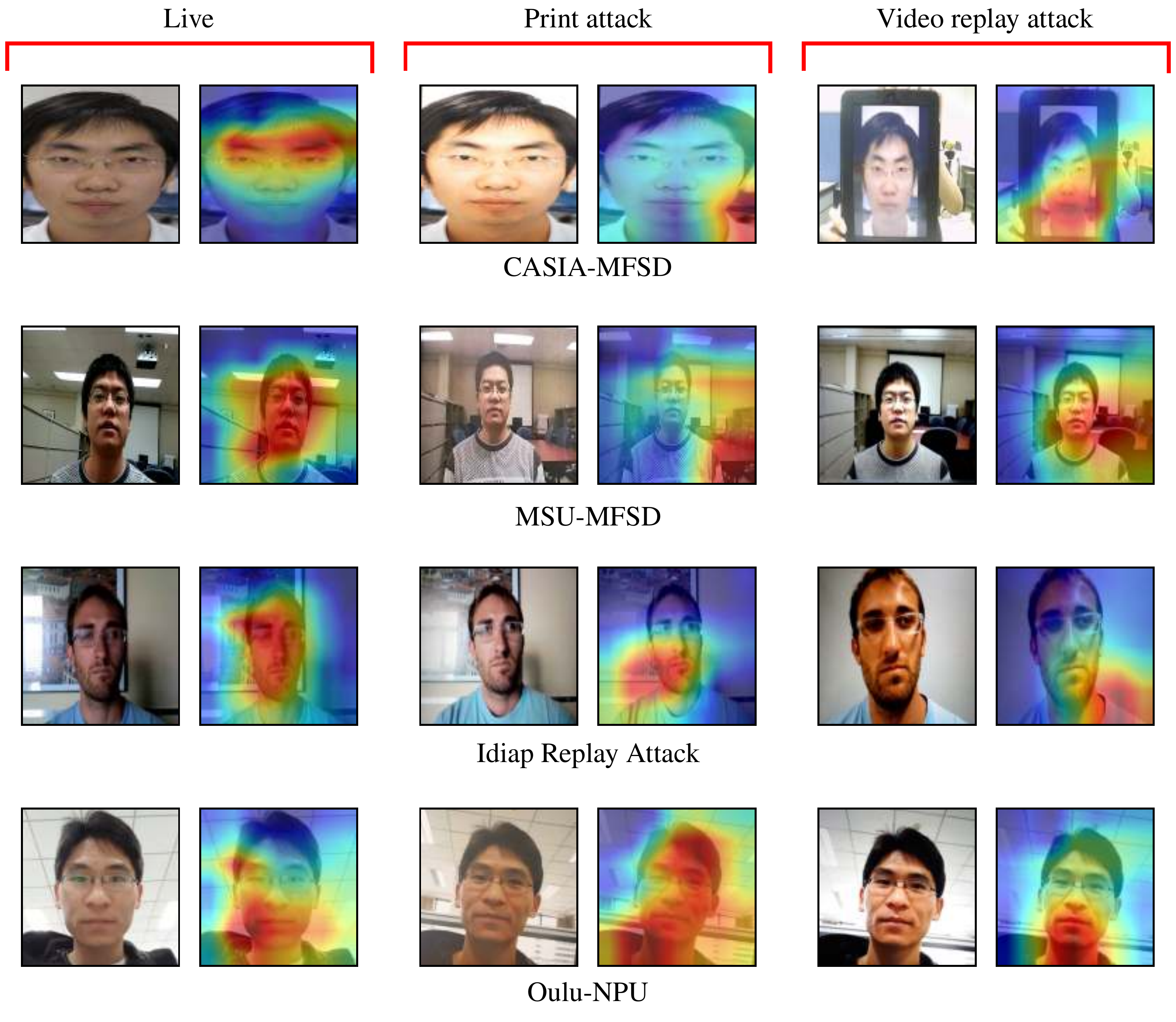} 
\end{center}
  \caption{Class activation map visualization of the proposed network. 
   For each column i.e. live, print attack and video replay attack, the original input images and their associated network class activation maps are shown.}
\label{fig:gradcam}
\end{figure*}

\subsection{t-SNE Visualization of the learned CNN features}

We compared the t-SNE visualization (Section 4.4, Fig.4 in the main)
of the CNN activations coming from the ResNet-50 baseline vs our proposed model
which were trained and tested according to the domain generalization setup - \emph{O\&C\&M$\to$I} (see Table \ref{table:dg_dataset}).
In Fig. \ref{fig:tsne2}, we present a t-SNE visualization for the ResNet-50 vs our proposed model
trained on the remaining three domain generalization training sets -
\emph{O\&C\&I$\to$M}, \emph{O\&M\&I$\to$C} and \emph{I\&C\&M$\to$O} (see Table \ref{table:dg_dataset}).
Each row in Fig. \ref{fig:tsne2} represents a domain generalization train/test setup (see Table \ref{table:dg_dataset}).
The plots in the first column (in Fig. \ref{fig:tsne2}) are generated using ResNet-50 baseline model. 
Whereas, the plots in the second and third columns (in Fig. \ref{fig:tsne2}) are generated using our proposed model.
For the sake of better visualization, however, we have deactivated the visualization of the target domain in the second column.

Similar observations can be made as in Fig.4 in the main paper.
The t-SNE plots in the second and third columns show that our model
(1) learns more discriminative features for live and spoof images (second column);
(2) aligns well the target domain’s live and spoof features with the source domains' live and spoof features.
In contrast, the ResNet-50 features show relatively weaker generalization ability on the target domain, as shown in Fig. \ref{fig:tsne2} (first column).

\subsection{Class activation map visualization}
Similar to Section 4.6 in the main paper,
here we present some additional class activation maps
for the ``live'', ``print attack'' and ``replay attack'' samples from the four face anti-spoofing datasets.
Fig. \ref{fig:gradcam} shows the class activation maps which are generated using Grad-CAM~\cite{selvaraju2017grad}.
Similar observations can be made (as in Section 4.6 in the main paper)
from these activation maps, i.e. 
for ``live'' samples, the network activations are high around the facial regions.
For ``print attacks'', the network activations are high in the background regions (except Oulu-NPU), i.e.
the network learns to classify print attacks by detecting the small artifacts often appears on the surface of the  paper material (on which the face image was printed).
The high resolution print attacks of Oulu-NPU might force the network to look at both facial regions as well as the background.
For ``video replay attack'', the network tries to gather information both from the facial regions and background.
The important clues to classify a replay attack might include the moire patterns appears in the CRT displays, the unique texture of the display screen frame etc.